\documentclass{article}

\usepackage[utf8]{inputenc}
\usepackage[T1]{fontenc}
\usepackage{hyperref}
\usepackage{url}
\usepackage{booktabs}
\usepackage{amsfonts}
\usepackage{nicefrac} 
\usepackage{microtype}

\usepackage{spconf,amsmath,amssymb,graphicx}
\usepackage{adjustbox} 

\usepackage{booktabs}
\usepackage{wrapfig}
\usepackage{array}
\usepackage{tablefootnote}
\usepackage{color}
\usepackage{cite}

\usepackage{wrapfig}

\usepackage{multirow,mathtools } \usepackage{algorithm,algpseudocode}

\usepackage{pifont}
\usepackage{color, colortbl}

\usepackage{blindtext}
\usepackage{lipsum}
\usepackage{threeparttable}

\usepackage{amsmath,amsfonts,bm}

\def\eqref#1{(\ref{#1})}

\def\1{\bm{1}}

\DeclareMathAlphabet{\mathsfit}{\encodingdefault}{\sfdefault}{m}{sl}
\SetMathAlphabet{\mathsfit}{bold}{\encodingdefault}{\sfdefault}{bx}{n}

\DeclareMathOperator*{\argmin}{arg\,min}

\DeclareMathOperator*{\minimize}{\text{minimize}}

\DeclareMathOperator*{\st}{\text{subject to}}

\newcommand{\btheta}{\boldsymbol{\theta}}

\newcommand{\bdelta}{\boldsymbol\delta}

\newcommand{\Def}[0]{\mathrel{\mathop:}=}

\usepackage{algorithm,algpseudocode}
\algnewcommand{\algorithmicforeach}{\textbf{for each}}
\algdef{SE}[FOR]{ForEach}{EndForEach}[1]
  {\algorithmicforeach\ #1\ \algorithmicdo}
  {\algorithmicend\ \algorithmicforeach}

\usepackage{pifont}

\usepackage{color, colortbl}
\definecolor{Gray}{gray}{0.93}
\definecolor{Orange}{rgb}{1,0.5,0}
\definecolor{DGray}{gray}{0.83}
\definecolor{LightCyan}{rgb}{0.88,1,1}

\usepackage[most]{tcolorbox}

\newcommand{\bbE}{\mathbb{E}}
\newcommand{\RR}{\mathbb{R}}
\newcommand{\sx}{\mathsf{x}}
\newcommand{\sy}{\mathsf{y}}
\newcommand{\bw}{\mathbf{w}}
\newcommand{\bx}{\mathbf{x}}

\newcommand{\bP}{\mathbf{P}}
\newcommand{\cD}{\mathcal{D}}
\newcommand{\cF}{\mathcal{F}}
\newcommand{\cS}{\mathcal{S}}
\newcommand{\cT}{\mathcal{T}}
\newcommand{\cX}{\mathcal{X}}
\newcommand{\cY}{\mathcal{Y}}

\newcommand{\ours}{\textsc{DERPLL}}

\title{
Robustness-preserving Lifelong Learning via Dataset Condensation
}
\name{Jinghan Jia$^{1}$, Yihua Zhang$^{1}$, 	Dogyoon Song$^2$, Sijia Liu$^1$, Alfred Hero$^2$}
\address{
$^1$ Dept. CSE, Michigan State University, MI 48824, USA \\
$^2$Dept. EECS, University of Michigan, Ann Arbor, MI 48109, USA
}

\begin{document}

\maketitle

\begin{abstract}
Lifelong learning (LL) aims to improve a predictive model as the data source evolves continuously. 
Most work in this learning paradigm has focused on
resolving the problem of `catastrophic forgetting,' which refers to a notorious dilemma between
improving model accuracy over new data and retaining accuracy over previous data. 
Yet, 
it is also known that machine learning (ML) models can be vulnerable in the sense that even tiny, adversarial input
perturbations can deceive the models into producing erroneous predictions. 
This motivates the research 
  objective of   this paper -- specification of a new LL framework that can salvage model robustness (against adversarial attacks) from catastrophic forgetting. Specifically, we propose a new memory-replay LL strategy
  that leverages modern bi-level optimization techniques to determine the `coreset' of the current data (\textit{i.e.}, a small amount of data to be memorized) for ease of preserving adversarial robustness over time. We term the resulting LL framework   `Data-Efficient Robustness-Preserving LL' ({\ours}).
The effectiveness of {\ours} is evaluated for class-incremental image classification using ResNet-18 over the CIFAR-10  dataset. Experimental results show that {\ours} outperforms the conventional coreset-guided LL baseline and achieves a substantial improvement in  both standard accuracy and robust accuracy. 

\end{abstract}


\section{Introduction}
\label{sec:intro}
While ML technologies have achieved significant advances for many use cases \cite{krizhevsky2017imagenet, devlin2018bert,zhou2020graph}, the problem of continual learning remains largely open. 
In continual, lifelong learning, ML models keep updated as new data come in or as the prediction task evolves over time. For example, 
the number of classes evolves over time in the 
\textit{class-incremental learning (CIL)} setting \cite{thrun1995learning, borsos2020coresets, chenqueried}, 
which is the setting considered in this paper.
The so-called \textit{catastrophic forgetting} phenomenon \cite{mccloskey1989catastrophic,  goodfellow2013empirical} occurs when past knowledge is dominated by newly acquired information, and it is a major obstacle to accurate continual learning. 
This paper develops a strategy to  mitigate the forgetting effects and maintain high performance  over tasks in the LL setting.

To alleviate catastrophic forgetting, numerous methods have been proposed and utilized in LL or CIL \cite{lopez2017gradient,  he2018exemplar, chrysakis2020online, zhang2020class,chenqueried,kirkpatrick2017overcoming,li2017learning, javed2018revisiting }; 
these methods principally rely on two types of techniques, namely, \emph{memory replay} and \emph{regularization}.
Firstly, memory replay \cite{lopez2017gradient,  he2018exemplar, chrysakis2020online, zhang2020class,chenqueried} stores a small dataset -- either by extracting a subset of the past training set \cite{lopez2017gradient,  chrysakis2020online} or by generating a synthetic summary \cite{he2018exemplar} -- to retain useful knowledge for the previous tasks, and then reuses it to `remind' the model of the old tasks when the model is being updated for new tasks.
Secondly, regularization techniques \cite{kirkpatrick2017overcoming,li2017learning, javed2018revisiting} have been successful in enforcing the newly updated model to prevent deviations far from the old model. 
In practice, relay and regularization techniques are used in combination.

There has been a recent surge in interest in \emph{adversarial robustness} of ML models to overcome their well-known fragility, especially those having deep architectures \cite{Goodfellow2014explaining, croce2020reliable}; ML models can be deceived by a tiny amount of (adversarial) perturbations on the input induced by an adversarial attacker.
While the past decade has witnessed significant progress  in adversarially robust ML in the static learning setting \cite{madry2017towards, zhang2022revisiting},  adversarial robustness for LL is largely unexplored.
Chen \textit{et at.} \cite{chenqueried} attempted to achieve adversarially robust CIL by applying adversarial training techniques to the streaming data setting. 
However, extra side information (\textit{i.e.}, queried unlabeled data) have to be assumed for addressing the problem of  `catastrophic forgetting'   \cite{chenqueried}. This may cause a new problem in data efficiency. 


In this work, we propose a novel approach to sustain adversarial robustness for models learned in the CIL setting in a data-efficient manner. 
Specifically, our strategy combines a memory-replay method based on coreset selection and a regularization technique to retain both prediction accuracy and adversarial robustness. 
We find that a carefully curated data summary is substantially more effective than a random subset for maintaining accuracy and adversarial robustness.


\vspace*{1mm}
\noindent \textbf{Contributions.} We summarize our contributions below. 

\noindent 
(1) We identify the outstanding challenges in attaining adversarially robust CIL. 

\noindent 
(2) We propose the {\ours} strategy to address `robustness forgetting' based on bi-level optimization implementation of coreset selection.

\noindent 
(3) We demonstrate the effectiveness of {\ours} for CIL  with deep models trained on real-world  image datasets.

\section{Background \& Related Work}
\vspace*{-1mm}
\textbf{Class-Incremental Learning (CIL).} 
In the LL paradigm, this paper will focus on 
CIL, which refers to the setting where a model is learned to solve a time-evolving task using a stream of data from a non-stationary distribution \cite{thrun1995learning, masana2020class}. 
CIL has received increasing attention in LL \cite{lopez2017gradient,  he2018exemplar, chrysakis2020online, zhang2020class,chenqueried,kirkpatrick2017overcoming,li2017learning, javed2018revisiting}, and various approaches have been proposed to alleviate the \emph{catastrophic forgetting} phenomenon (\textit{i.e.},   the interference to the previously learned knowledge caused by acquiring new information) \cite{mccloskey1989catastrophic, goodfellow2013empirical}. 
The CIL methods devised to mitigate the forgetting effects largely rely on two types of techniques, namely, \emph{memory replay} and \emph{regularization}.
The \emph{memory-replay}-type approaches maintain a small number of data points to `memorize' the old tasks and replay them during the learning phase for new tasks.
To construct a summarizing dataset, some of these methods directly extract a subset from the past training dataset \cite{lopez2017gradient,  chrysakis2020online}, while others generate an artificial summary for the old tasks using generative models \cite{he2018exemplar}, unlabeled auxiliary data \cite{zhang2020class,chenqueried}, or knowledge distillation \cite{javed2018revisiting, chenqueried}. 
Another category of widely used CIL methods utilizes regularization techniques to prevent the newly trained model from deviating too far from the previously trained models \cite{kirkpatrick2017overcoming,li2017learning}. 
Learning without forgetting (\emph{LwF}), coined by Li and Hoiem \cite{li2017learning}, is the epitome of this type of approach, which utilizes ideas from knowledge distillation to restrain forgetting. LwF and its variants form a popular and successful family of regularization methods for CIL \cite{li2017learning,javed2018revisiting, chenqueried}.

\vspace*{1mm}
\noindent \textbf{Coreset Selection.} 
The coreset selection problem has a long history in computational geometry \cite{agarwal2005geometric}, and has appeared in various contexts as a tool for dimensionality reduction \cite{feldman2020turning}. Applications of coreset selection in ML include $k$-means clustering \cite{feldman2011unified}, logistic regression \cite{huggins2016coresets}, Gaussian mixture model learning \cite{lucic2017training}, active learning \cite{sener2018active}, and Bayesian posterior inference \cite{campbell2019automated} and relational data analysis \cite{curtin2020rk}. 
Recently, Borsos \emph{et al.} \cite{borsos2020coresets} proposed a coreset-selection-based approach to maintain the high accuracy of the models in the task of continual learning. 
They reformulated the coreset selection problem as a cardinality-constrained bi-level optimization problem and proposed a greedy approach. 
However, this greedy method is difficult to scale due to its large computational overhead.

\vspace*{1mm}
\noindent \textbf{Adversarial Training (AT).}
Recognition of the fragility of ML models has promoted a flurry of research efforts to devise effective defenses against adversarial attacks \cite{madry2017towards,zhang2019theoretically,zhang2019you}. 
In particular, adversarial training (AT) \cite{madry2017towards}, which trains a model by minimizing the worst-case (maximum) training loss, has been recognized as one of the most effective defense methods \cite{athalye2018obfuscated}. 
Although adversarial robustness has been extensively studied via AT or its variants 
\cite{zhang2019theoretically,zhang2019you}, these efforts are mostly restricted to robustifying ML models in the \textit{static} learning paradigm. 
By contrast, the adversarial robustness of LL is little  explored, except \cite{chenqueried}  to the best of our knowledge. In \cite{chenqueried}, the problem of  \emph{robustness forgetting} (\textit{i.e.},  catastrophic forgetting of of adversarial robustness over time-evolving tasks) was identified in   CIL. The authors   attempted to tackle this challenge using robustified regularization techniques, \textit{e.g.}, TRADES \cite{zhang2019theoretically}, with the assistance of auxiliary unlabeled data. 
Compared with \cite{chenqueried},  we revisit the problem of adversarially robust CIL and propose a  data-efficient robustness-preserving CIL framework, which
 does not  require additional side information to mitigate the catastrophic forgetting of adversarial robustness.
\vspace*{-3mm}

\section{Problem: Adversarially Robust CIL}
\label{sec: preliminaries}
In this section, we formally define the problem setting we consider, and identify the outstanding challenges in sustaining the adversarial robustness of models for CIL. 
To this end, we start from a basic formulation for learning problems, and gradually introduce additional layers of complexity. 

Consider a random vector $(\sx, \sy) \in \cX \times \cY$, which is drawn from an unknown joint distribution $\bP$. 
We refer to $\sx, \sy$ as input, and response (or, label), respectively; we commonly have $\cX = \RR^d$; and 
$\cY = \{ 1, \dots, K \}$ (classification). 
We have access to a finite-sized sample $\cD = \{ (\bx_i, y_i): i=1, \dots, n \}$ drawn from $\bP$. 
Let $\cF$ denote the set of functions we consider, \textit{e.g.}, those implementable by a DNN (deep neural network), parameterized by model parameters $\btheta$. 
The goal of ML is to find a function $f_{\btheta} \in \cF$ that approximates well the conditional expectation of $\sy$ given $\sx$, \textit{i.e.}, $f_{\btheta}(\bx) \approx \bbE_{\bP} [ \sy| \sx = \bx ]$,  using the finite dataset $\cD$.

\vspace*{1mm}
\noindent \textbf{Class-incremental Learning (CIL).}
Observe from above that the unknown underlying distribution $\bP$ along with a dataset $\cD$ defines the target `task' of interest.  CIL refers to the setting where $\bP$ evolves over time, \textit{e.g.}, the number of class labels increases. 
We model this CIL setup using a discrete-time task sequence $( \cT^{(t)} : t=1, 2, \dots )$, where the task at time $t$ is defined as a pair $\cT^{(t)} = ( \cD^{(t)}, \bP^{(t)} )$ of dataset and (unknown) distribution.
When faced with a stream of tasks as above, the goal of CIL is to learn a sequence of models $( \btheta^{(t)}: t=1, 2, \dots )$ such that $\btheta^{(t)}$ solves task $\cT^{(t)}$, and at the same time, retains good performance for all previous tasks $\cT^{(1)}, \dots, \cT^{(t-1)}$.

\vspace*{1mm}
\noindent \textbf{Adversarially Robust Learning.}
The goal of adversarially robust learning is to acquire a robust model $\btheta$ against adversarial input perturbations. For instance, adversarial training (AT) attempts to learn $\btheta$ by solving \cite{madry2017towards}

\vspace*{-5mm}
{\small \begin{align}
    \begin{array}{l}
      \displaystyle  \min_{\btheta}     ~ \ell^{\epsilon}_{\mathrm{AT}}(\btheta; \cD) \Def  \bbE_{(\bx,y) \in \cD} \Big\{ \max_{\| \boldsymbol \delta\|_\infty \leq \epsilon} \ell (\btheta; \bx + \bdelta, y) \Big\} ,
    \end{array}
    \label{eq: AT_prob}
    \tag{AT}
\end{align}}%
where $\boldsymbol \delta$ is a variable representing the input perturbation that falls in the $\ell_\infty$-norm ball of radius $\epsilon$, and $\ell$ is a loss function in use for training. 
That is, AT `robustifies' the learned model by modifying the loss used in the standard empirical risk minimization formulation to reflect the worst-case behavior; $\ell(\btheta; \bx, y) \mapsto \max_{\| \bdelta \|_\infty \leq \epsilon} \ell( \btheta; \bx + \bdelta, y)$.

\vspace*{2mm}
\noindent \textbf{Main Problem: Challenges in Adversarially Robust CIL.}
Both CIL and adversarial robustness capture important features in practical ML problems. However, these two topics have been studied   separately in the literature. In this work, {we aim to} address adversarial robustness in the CIL setting. 
Specifically, we focus on two challenges that arise while we attempt to na\"ively extend AT to CIL:


\textbf{(C1) Lack of principles for memory replay:} 
While memory replay is a popular approach for CIL, it is unclear what data should be stored to sustain both accuracy and robustness. 

\textbf{(C2) Forgetting of robustness:} 
CIL may fail to  retain adversarial robustness over a sequence of time-evolving tasks. 

Inspired by these challenges, we ask the question:
 \begin{tcolorbox}[left=1.2pt,right=1.2pt,top=1.2pt,bottom=1.2pt]
 \begin{center}
\textbf{(P)}: \textit{How to  design 
a new CIL paradigm that can optimize  data efficiency and model robustness  simultaneously?}
\end{center}
\end{tcolorbox}

\section{Methods}
\label{sec: Methods}
In this section, we describe our {\ours} approach to tackle \textbf{(P)}, which combines a memory-replay technique based on coreset selection and a LwF-type regularization technique.
\vspace*{-3mm} 
\subsection{Coreset Selection for Memory Replay}\label{sec:coreset}
Coreset selection aims at `compressing' the dataset $\cD := \bigcup_{k=1}^{t-1} \cD^{(k)}$ into a small summary $\cS^{(t)}$ that contains sufficient information of previous tasks $\cT_1, \dots, \cT_{t-1}$ for training $\btheta^{(t)}$. 
The coreset selection problem can be formulated as the following bi-level optimization (BLO) problem \cite{borsos2020coresets}:

\vspace*{-5mm}
{\small \begin{align}
    \begin{array}{cl}
    \displaystyle \minimize_{\mathbf w \in \Omega}     & \displaystyle \frac{1}{N}\sum_{i=1}^N \ell \big(\btheta^*(\mathbf w); \mathbf x_i, y_i \big) \\
    \st      &  \btheta^*(\mathbf w) = \displaystyle \argmin_{\btheta} \frac{1}{N}\sum_{i=1}^N   w_i \cdot \ell(\btheta; \mathbf x_i, y_i) ,
    \end{array}
    \hspace*{-5mm}
    \label{eq: coreset_ori}
\end{align}}%
where $\mathbf w = [w_1, \ldots, w_N]^T \in \{0, 1\}^N$ is the selection variable, $\Omega = \{ \mathbf w \in \{ 0, 1\}^N \, | \, \mathbf 1^T \mathbf  w = n \}$ encodes the cardinality constraint (\textit{i.e.}, the size of the coreset), and $\ell$ is a model training loss (\textit{e.g.}, cross entropy loss).
Note that problem \eqref{eq: coreset_ori} involves two levels of entangled optimization sub-problems: (i) the \textit{lower-level} problem (\textit{w.r.t.} $\btheta$) finds the best model given a coreset given the upper-level variables $\mathbf w$; (ii) the \textit{upper-level} problem finds the best coreset (\textit{i.e.}, the best $\bf w$) by assessing the performance of the model $\btheta^*(\mathbf w)$ over the entire dataset.

\vspace{1mm}
\noindent \textbf{Computational Challenge in Solving \eqref{eq: coreset_ori}.}
While \eqref{eq: coreset_ori} provides a concise formulation, it can be difficult to solve for two reasons: 
(i) the cardinality constraint on the Boolean variable $\bf w$; and 
(ii) the need of hard implicit gradient computation $\frac{d \btheta^*(\mathbf w)}{d \mathbf w}$ to obtain a closed-form expression for the upper-level gradient.
To circumvent these computational challenges, Borsos \emph{et al.} \cite{borsos2020coresets} developed a greedy algorithm that constructs a coreset by incrementally adding a training sample with the maximum influence score (IS) -- \textit{i.e.}, the sensitivity of the upper-level objective with respect to the change in $w_i$. 
While it is possible to compute the closed form of IS using the influence function theory \cite{cook1982residuals,koh2017understanding}, it involves the inverse Hessian of the inner objective w.r.t. $\btheta$. This makes the use of IS impractical for deep models such as ResNet-18 \cite{he2016deep} used in our experiments.

\vspace{1mm}
\noindent \textbf{Our Approaches to Solve \eqref{eq: coreset_ori}.}
We take an alternative approach to numerically solve \eqref{eq: coreset_ori}. 
To be precise, we adopt two techniques: 
\textcolor{blue}{(i)} we relax the Boolean constraint $\bw \in \{0,1\}^N$ to a more amenable box constraint $\bw \in [0, 1]^N$, enabling the use of standard BLO solvers; and 
\textcolor{blue}{(ii)} we leverage a \textit{straight-through (ST)} gradient estimator \cite{bengio2013estimating}, which retains the discrete function value evaluation in the forward pass of DL but the continual derivative evaluation in the backward pass.  
Concretely, we reformulate \eqref{eq: coreset_ori} as the following:

\vspace*{-5mm}
{\small \begin{align}
    \begin{array}{cl}
    \displaystyle  \minimize_{\bw \in {\color{blue} \Omega^\prime} }     &\displaystyle   f(\bw) \Def  \frac{1}{N}\sum_{i=1}^N \ell \big(\btheta^*(  \bw  ); \bx_i, y_i \big)  \\
    \st      &  \btheta^*(\bw) = \displaystyle \argmin_{\btheta} \frac{1}{N}\sum_{i=1}^N {\color{blue} \mathcal B_i( \mathbf{w}) } \cdot \ell(\btheta; \mathbf x_i, y_i),
    \end{array}
    \hspace*{-5mm}
    \label{eq: coreset_ori_relax}
\end{align}}%
where $\Omega^\prime = \{ \bw \in [  0, 1 ]^N \, | \, \mathbf 1^T \bw = n\}$, and $\mathcal B_i: \RR^N \to \{0,1\}$ is a thresholding function such that $\mathcal B_i ( \bw ) = 1$ if and only if $w_i \geq w_{[n]}$ where  $w_{[n]}$ is the $n$-th largest element of $\bw$.
\vspace*{-1mm}

The reformulated problem \eqref{eq: coreset_ori_relax} bypasses the two challenges in \eqref{eq: coreset_ori}. 
The upper-level problem is continuous, and the projected gradient descent (PGD) method is applicable. 
In addition, the derivative   of thresholding functions in the ST gradient estimator is omitted and thus makes the computation of implicit gradient $\frac{d \btheta^*(\bw)}{d \bw}$ tractable in the continuous domain.  Thus, we can utilize standard DL toolkits to compute the implicit gradient, \textit{e.g.},
gradient unrolling \cite{liu2021investigating},
which uses automatic differentiation to compute the implicit gradient evaluated at a finite-step gradient-descent approximation of $\btheta^*(\bw)$.

\vspace*{-3mm}
  
\subsection{{\ours}: Robustness-aware LL with Coresets}
Next, we illustrate how to integrate the selected coreset $\cS^{(t)}$ to achieve data-efficient robustness-preserving lifelong learning ({\ours}). 
As described in Sect.\,\ref{sec: preliminaries}, we aim to sustain good performance of the ML models $\btheta^{(t)}$ over continually evolving tasks $\cT^{(t)}$. 
We propose to train $\btheta^{(t)}$ using $\cD^{(t)}$ and $\cS^{(t)}$, with a training objective function that is designed to meet the desiderata -- \textit{i.e.}, simultaneous adaptation and preservation.

To this end, we design an objective function that consists of two components. 
First, we utilize the usual \ref{eq: AT_prob} loss, \textit{i.e.}, $\ell^{\epsilon}_{\mathrm{AT}}(\btheta^{(t)}; \cD^{(t)})$ to ensure robustness at the current time $t$. 
Second, we introduce an LwF loss term to retain accuracy/robustness on the past tasks, which is implemented by minimizing the prediction disagreement between the old model $\btheta^{(t-1)}$ and the new model $\btheta^{(t)}$ on the archived coreset, $\cS^{(t)}$: 

\vspace*{-5mm}
{\small
\begin{align}
  &  \ell_{\mathrm{LwF}} (\btheta^{(t)}; \btheta^{(t-1)}, \cS^{(t)}) 
    = 
    \bbE_{ \bx \in \cS^{(t)}}  \big\{     d (\btheta^{(t)}, \btheta^{(t-1)}; \bx) \big\}
     \nonumber \\
   & \hspace*{17mm}  +
     \bbE_{ \bx \in \cS^{(t)}}  \Big\{   \max_{\| \bdelta \|_{\infty} \leq \epsilon }  d (\btheta^{(t)}, \btheta^{(t-1)}; \bx + \bdelta) \Big\},
   \nonumber 
\end{align}}%
where $d (\btheta_1, \btheta_2; \bx )$ is a disparity measure (\textit{e.g.}, KL divergence \cite{chenqueried} used in our experiments) of predictions from two models $\btheta_1$ and $\btheta_2$ at input $\bx$; the two terms are designed to penalize forgetting of accuracy and robustness, respectively.

Overall, we propose 
the {\ours} framework that learns $\btheta^{(t)}$ by minimizing the composite objective function, \textit{i.e.}, $\ell^{\epsilon}_{\mathrm{AT}}$ regularized with $\ell_{\mathrm{LwF}}$:

\vspace*{-3.9mm}
{\small
\begin{align}
\displaystyle \minimize_{\btheta^{(t)}} ~  \ell^{\epsilon}_{\mathrm{AT}} (\btheta^{(t)}; \mathcal D^{(t)}) + \gamma   \ell_{\mathrm{LwF}} (\btheta^{(t)}; \btheta^{(t-1)}, \cS^{(t)}),
\label{eq: CIL_Coreset}
\end{align}}%
where $\gamma > 0$ is a tunable parameter. After acquiring $\btheta^{(t)}$, the coreset $\cS^{(t)}$ is updated to $\cS^{(t+1)}$ by augmenting a newly selected subset from $\cD^{(t)}$; \textit{cf.} \eqref{eq: coreset_ori_relax} in Sect. \ref{sec:coreset}.

\section{Experiments}
\label{sec: experiments}
In this section, we evaluate the effectiveness of our proposal in the task of class-incremental image classification.

\vspace*{1mm}
\noindent \textbf{Datasets \& Model Architectures.} 
We consider the CIFAR-10  \cite{Krizhevsky2009learning} dataset  using the neural network model ResNet-18 \cite{he2016deep}. In the context of CIL, CIFAR-10 with 10 classes is randomly divided into 5 tasks, each with 2 classes. The model is trained with a time-evolving task stream and will update its classification ability     when a new task arrives. 

\vspace*{1mm}
\noindent \textbf{Implementation Details.} For each new task, the model will be adversarially trained with both the \textit{complete} new training data and the \textit{selected} old data from the memory bank stored from previous tasks. The coreset selection for the new data will take place after the model is updated at the current task. We set the size of the memory bank to 100 images per class if not otherwise specified. We leverage the widely adopted 10-step PGD attack \cite{madry2017towards} to generate adversarial examples with the attack strength of $\epsilon=8/255$, where $\epsilon$ was   defined in \eqref{eq: AT_prob}.
To solve problem \eqref{eq: CIL_Coreset}, we use the SGD optimizer as the model training recipe. And we tune the regularization parameter $\gamma $ in $[10^{-3}, 1]$, and find that $\gamma = 0.1$   consistently provides the best  robustness and accuracy performance over time.

\vspace*{1mm}
\noindent \textbf{Baselines \& Evaluation Metrics.} We consider two data selection oracles as our baselines:
\textbf{\ding{182} random selection} randomly samples a subset of the memory bank size from the training set; \textbf{\ding{183} influence score-based selection (IS) \cite{borsos2020coresets}} selects the data with the top influence scores. For evaluation, we evaluate the performance on \textit{all the tasks} encountered so far. In particular, we look into the \textbf{\ding{172} standard accuracy (SA)}, the test accuracy on the clean test set; and \textbf{\ding{173} robust accuracy (RA)}, tested on the perturbed test set. We perform 20-step PGD attack to generate test-time perturbations for evaluation with the same attack strength ($\epsilon = 8/255$) as training.


\begin{table}[htb]
\begin{center}
\vspace*{-1mm}
\caption{\small{Performance comparison  of different methods for robustness-aware CIL under (CIFAR-10, ResNet18). 
Each row shows   the standard accuracy (SA) and robust accuracy (RA) on  old tasks and the current task. For example, in the first row, $\mathcal T^{(1)}$ is the old task, and $ \mathcal T^{(2)}$ is the current task.
The best results are   in \textbf{bold}.}}
\label{tab: robustness}
\begin{threeparttable}
\resizebox{1\linewidth}{!}{
\begin{tabular}{c|c|cc|cc|cc|cc|cc}
\toprule
\multirow{3}{*}{Tasks} & \multirow{3}{*}{Methods} & \multicolumn{8}{c}{CIFAR-10 (SA, RA)} \\ \cmidrule{3-12} 
& & \multicolumn{2}{c|}{$\mathcal{T}^{(1)}$ (\%)} & \multicolumn{2}{c|}{$\mathcal{T}^{(2)}$ (\%)} & \multicolumn{2}{c|}{$\mathcal{T}^{(3)}$ (\%)} & \multicolumn{2}{c|}{$\mathcal{T}^{(4)}$ (\%)} & \multicolumn{2}{c}{$\mathcal{T}^{(5)}$ (\%)} \\ \cmidrule{3-12} 

& & \multicolumn{1}{c|}{SA} & \multicolumn{1}{c|}{RA} & \multicolumn{1}{c|}{SA} & \multicolumn{1}{c|}{RA} & \multicolumn{1}{c|}{SA} & \multicolumn{1}{c|}{RA} & \multicolumn{1}{c|}{SA} & \multicolumn{1}{c|}{RA} & \multicolumn{1}{c|}{SA} & \multicolumn{1}{c}{RA} 

\\\midrule
\multirow{3}{*}{$\mathcal{T}^{(1)}\sim\mathcal{T}^{(2)}$}
& Random & 83.20 & 65.30& 95.75 & 83.10
&- &-& -& -& -&- 
  \\

& IS & 84.40 & 66.00 & 97.00  & 83.75  
& - & - & - & - & - & - 
     \\

& Ours & \textbf{85.05} & \textbf{66.25} & \textbf{97.55} & \textbf{87.20}
& - & - & - & - & - & - 
   \\

\midrule
\multirow{3}{*}{$\mathcal{T}^{(1)}\sim\mathcal{T}^{(3)}$}
& Random & 80.80 & 62.85 & 93.35 & 79.30 & 94.10 & 84.90
& -& -& -&- 
\\

& IS & 81.75 &62.95 & 94.50 & 79.35 & 94.95 & 84.95
& - & - & - & - 
 \\

& Ours & \textbf{82.05} & \textbf{64.50} & \textbf{94.85} & \textbf{79.40} & \textbf{95.10} & \textbf{86.00}
& - & - & - & - 
 \\

\midrule
\multirow{3}{*}{$\mathcal{T}^{(1)}\sim\mathcal{T}^{(4)}$}
& Random & 75.85 & 59.90 & 82.80 & 63.10 & 89.40 & 76.65 & 75.90 & 62.70
& - & -  \\

& IS & 76.20 & 61.35 & 83.30 & 63.30 & 89.90 & 78.80 & 76.15 & 63.20
& - & -  \\

& Ours & \textbf{76.85} & \textbf{61.55} & \textbf{87.90} & \textbf{69.60} & \textbf{89.94} & \textbf{81.60}  & \textbf{77.00} & \textbf{63.25} 
& - & -  \\

\midrule
\multirow{3}{*}{$\mathcal{T}^{(1)}\sim\mathcal{T}^{(5)}$}
& Random & 71.85 & 55.00 & 77.60 & 60.10 & 84.40 & 72.05 & 74.60 & 55.60& 74.85 & 60.90 
	\\
& IS     & 73.05 & 55.40 & 79.15 & 61.65 & 85.40 & 72.35 & 75.35 & 56.10 & 76.60 & 62.20
  	\\
& Ours   & \textbf{74.10} & \textbf{58.20} & \textbf{84.35} & \textbf{63.70} & \textbf{87.75} & \textbf{75.35} & \textbf{76.40} & \textbf{56.80} & \textbf{78.20} & \textbf{64.50} 
  	\\
\bottomrule
\end{tabular}}
\end{threeparttable}
\end{center}
\vspace*{-7mm}
\end{table}


\vspace*{1mm}
\noindent \textbf{Experiment Results.} In Tab.\,\ref{tab: robustness}, we show the overview performance  of {\ours} and  baseline methods for robustness-aware CIL  on CIFAR-10 under ResNet-18.  We summarize our key observations below.
\textbf{First}, our method better helps  mitigate the catastrophic forgetting of both accuracy and robustness. 
Let us take  the row of `$\mathcal T^{(1)} \sim \mathcal T^{(4)}$'  as an example, where  $ \mathcal T^{(4)}$ is the   current task and   $ \mathcal T^{(i)}$ for $i \in \{ 1,2,3\}$ are   old tasks. We observe that our proposed method (\ours) outperform baselines in  model performance preservation on old tasks, as evidence by a substantial SA/RA improvement over all the baselines,  \textit{e.g.},  $4\%$ on SA and over $6\%$ on RA evaluated at $\mathcal T^{(2)}$ in the row of  `$\mathcal T^{(1)} \sim \mathcal T^{(4)}$'.
 \textbf{Second}, our method also achieves the best SA/RA at current tasks  across all time steps
  (corresponding to the diagonal of Tab.\,\ref{tab: robustness}). The above two observations indicate that  our method is effective not only in mitigating the forgetting of accuracy and robustness  but also in achieving high-accuracy prediction based on the new knowledge.  
\textbf{Third}, we note that all the methods suffer from a performance drop as more tasks are encountered. This is not surprising, since more tasks bring in more classes and thus raise greater challenges in adversarial training and  the mitigation of the forgetting issue.

\begin{wrapfigure}{r}{40mm}
\vspace*{-3mm}
\centerline{
\begin{tabular}{c}
\hspace*{-8mm}
\includegraphics[width=.25\textwidth,height=!]{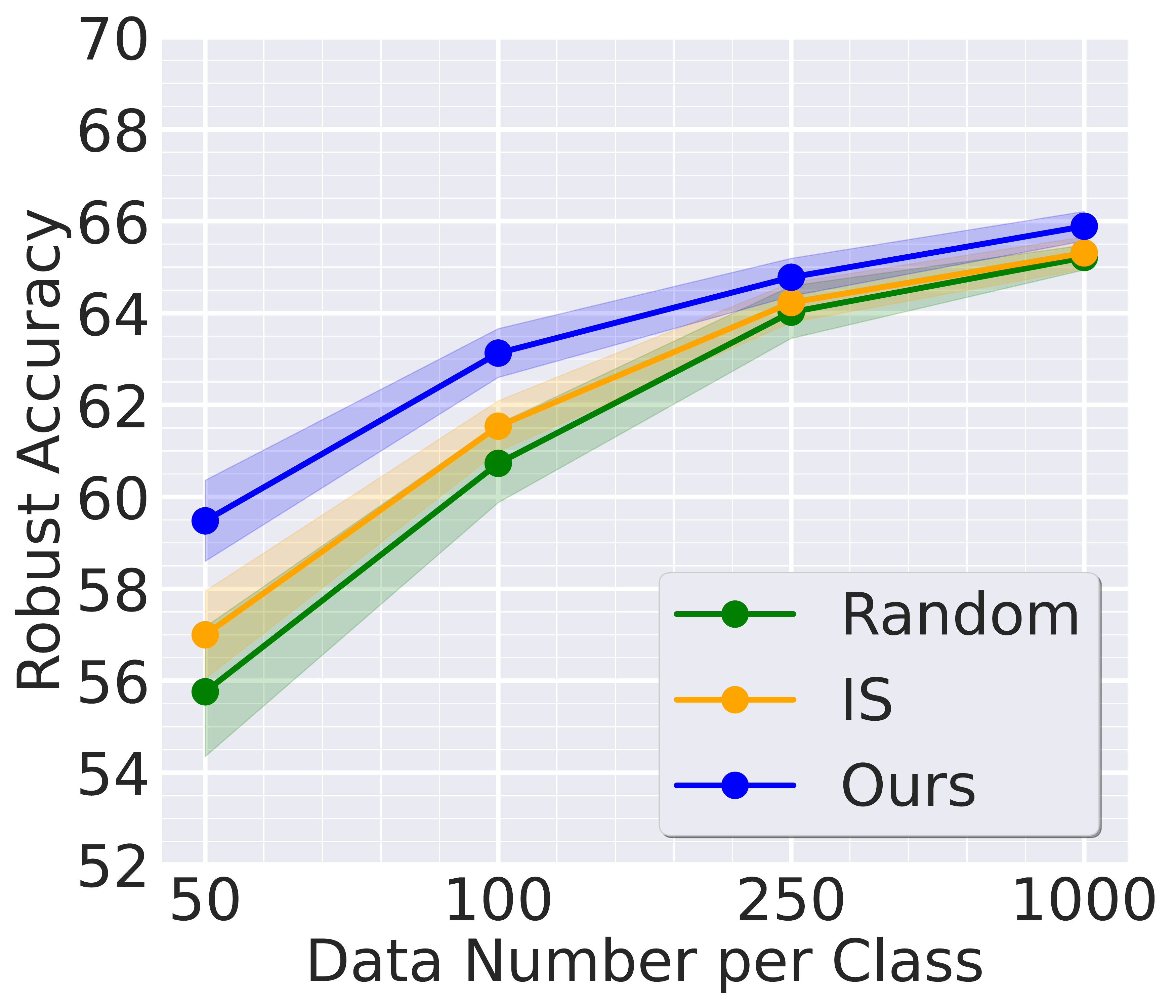} 
\end{tabular}}
\vspace*{-5mm}
    \caption{\small{RA of different methods \textit{v.s.} coreset sizes. 
    }} 
  \label{fig: rob_ratio}
\vspace*{-3.5mm}
\end{wrapfigure}
Fig.\,\ref{fig: rob_ratio} shows the model robustness (over the entire test set) vs. the coreset size (in terms of data number per class)  adopted  at a given time step, where the curve and shaded area represent the mean value and standard deviation over three independent trials.
We observe that the advantage of our method becomes more significant as the coreset size reduces. For example, compared to our default setting (100 images per class), the performance improvement over the \textsc{IS} baseline increases from $2\%$ to $3\%$. This is  encouraging when the data storage capacity is limited in CIL.


\section{Conclusion}
\vspace*{-2mm}
In this paper, 
we develop {\ours},  
the new data-efficient robustness-preserving lifelong learning framework. We design an advanced   coreset  selection method to determine the most effective data for ease of storing continual data while preserving model's adversarial robustness. Empirically, we show the effectiveness of  {\ours} in DNN-involved lifelong   learning for class-incremental image classification.

\clearpage
\newpage
{\small
\bibliographystyle{IEEEbib}
\bibliography{refs,bibs/ref_lifelong_SL,bibs/ref_SL_adv}
}
\end{document}